\documentclass[a4paper,fleqn]{cas-dc}

\usepackage[numbers]{natbib}

\usepackage{graphicx}
\graphicspath{{figures/}}

\usepackage{array}
\usepackage{booktabs}
\usepackage{caption}
\usepackage{subcaption}
\usepackage{amsmath,amssymb}
\usepackage{amsmath}
\newtheorem{prop}{Proposition}
\usepackage{graphicx}
\graphicspath{{figures/}}
\DeclareGraphicsExtensions{.jpg,.png,.pdf}

\usepackage{pgfplots}
\usepackage{pgfplotstable}
\usepackage{filecontents}
\usepackage{booktabs}
\usepackage{subcaption}
\usepackage{array}
\usepackage{multirow}
\usepackage{tabularx}
\newcolumntype{Y}{>{\centering\arraybackslash}X}
\newcolumntype{M}[1]{>{\centering\arraybackslash}m{#1}}

\usepackage{pifont}

\usepackage{algorithmic}
\usepackage[linesnumbered,ruled,vlined]{algorithm2e}
\SetKwInput{KwInput}{Input}  
\SetKwInput{KwOutput}{Output}

\usepackage[linesnumbered,ruled,vlined]{algorithm2e}

\usepackage{xurl}

\SetKwInput{KwInput}{Input}  
\SetKwInput{KwOutput}{Output}

\begin{document}

\let\WriteBookmarks\relax
\def\floatpagepagefraction{1}
\def\textpagefraction{.001}

\shorttitle{Using Anomaly Detection to Detect Poisoning Attacks in Federated Learning Applications}
\shortauthors{Ali Raza et~al.}

\title [mode = title]{Using Anomaly Detection to Detect Poisoning Attacks in Federated Learning Applications}

\author[1]{Ali~Raza}[orcid=0000-0001-8326-8325]
\cormark[1]
\author[2]{Shujun~Li}[orcid=0000-0001-5628-7328]
\author[3]{Kim~Phuc~Tran}[orcid=0000-0001-8326-8325]
\author[3]{Ludovic~Koehl}[orcid=0000-0002-3404-8462]
\author[4]{Kim~Duc~Tran}[orcid=0000-0002-4325-9528]

\address[1]{Honda Research Institute EU, Germany}

\address[2]{School of Computing \& Institute of Cyber Security for Society (iCSS), University of Kent, UK}

\address[3]{University of Lille, ENSAIT, GEMTEX–Laboratoire de Génie et Matériaux Textiles, F-59000 Lille, France}

\address[4]{International Research Institute for Artificial Intelligence and Data Science, Dong A University, Da Nang, Vietnam}

\cortext[cor1]{Corresponding author}

\nonumnote{The authors can be contacted via ali.raza@honda-ri.de (Ali Raza), kim-phuc.tran@ensait.fr (Kim Phuc Tran), ludovic.koehl@ensait.fr (Ludovic Koehl), hooklee@gmail.com or S.J.Li@kent.ac.uk (Shujun Li), ductk@donga.edu.vn (Kim Duc Tran).}

\begin{abstract}
Adversarial attacks such as poisoning attacks have attracted the attention of many machine learning researchers. Traditionally, poisoning attacks attempt to inject adversarial training data to manipulate the trained model. In federated learning (FL), data poisoning attacks can be generalized to model poisoning attacks, which cannot be detected by simpler methods due to the lack of access to local training data by the detector. State-of-the-art poisoning attack detection methods for FL have various weaknesses, e.g., the number of attackers has to be known or not high enough, working with i.i.d.\ data only, and high computational complexity. To overcome the above weaknesses, we propose a novel framework for detecting poisoning attacks in FL, which employs a reference model based on a public dataset and an auditor model to detect malicious updates. We implemented a detector based on the proposed framework and using a one-class support vector machine (OC-SVM), which reaches the lowest possible computational complexity $\mathcal{O}(K)$ where $K$ is the number of clients. We evaluated our detector's performance against state-of-the-art (SOTA) poisoning attacks for two typical applications of FL: electrocardiograph (ECG) classification and human activity recognition (HAR). Our experimental results validated the performance of our detector over other SOTA detection methods.
\end{abstract}


\maketitle

\section{Introduction}

Privacy and security are among the top issues to be addressed in privacy and security-sensitive applications of machine learning (ML), such as healthcare, autonomous vehicles, etc. Federated learning (FL) has been introduced to enhance data privacy in machine learning applications~\cite{mcmahan2016federated}. FL attempts to provide enhanced privacy to the data owners by collaboratively training a joint model by only sharing parameters of locally trained models, in this way the data owners never share raw data, but can collaboratively train a joint robust global model. For instance, various hospitals can train ML models jointly for healthcare applications without directly sharing their privacy-sensitive data. Generally speaking, FL iterates in three steps: the \textit{global server} (i.e., a cloud server), which maintains the \textit{global model}, sends the global model to the edge devices; the edge devices update the local models using their local training data and share the trained parameters of the locally trained model with the global server, and the global server updates the global model by incorporating the shared parameters according to an aggregation rule. For example, the mean aggregation rule, which computes the weighted average of the shared local model's parameters, is one of the widely used aggregation algorithms~\cite{sun2021}. Nevertheless, in such cases, the global model can be easily manipulated, even if a single-edge device is compromised~\cite{fang2020, blanchard2017, li2020}. The attack surface of FL is growing due to its distributed nature. For example, malicious peers can launch data poisoning~\cite{biggio2012, jagielski2018} or model poisoning~\cite{zhou2021} attacks, in which one or more malicious edge devices manipulate their local training data or the local model trained on benign data, to impair the performance of the updated global model.

FL can be divided into three phases: data and behavior auditing, training, and testing. FL faces different kinds of security threats in each phase~\cite{liu2022threats}. Hence, establishing secure FL needs to take effective measures at each phase to mitigate such threats. A solution before integrating a local model into the global model is to audit the local data before training the local model(s). However, due to privacy concerns and the architecture of FL, it is challenging to conduct such audits~\cite{liu2022threats}. A trivial method to address model poisoning attacks could be using accuracy, i.e., using accuracy as a measure to examine the quality of data being used to train the local model. Nevertheless, such methods cannot be generalized as accuracy alone cannot reveal information about the underlying data. Just looking at the accuracy it is hard to claim that a local model is trained on benign or malicious data. Furthermore, models can be designed to have high accuracy for the testing samples by including them in the training dataset. A model can have low accuracy even if it has been trained on benign data depending on the amount of training data, training epochs, hyper-parameters tuning, optimization, etc. Hence, new solutions are required to detect such model and data poisoning attacks. Methods should be developed to verify that the shared local model gradients are not trained on anomalous or malicious (e.g., featured poisoning, label poisoning) data. In other words, malicious behaviors of local models should be detected before considering them in the global aggregation process to prevent malicious peers from compromising and manipulating the global model. Therefore, many researchers have proposed different poisoning attack detection methods for FL~\cite{Chen2021, shafahi2018, Xia2023}, which unfortunately all suffer from various weaknesses, e.g., the number of attackers has to be known or cannot be too high, working with i.i.d.\ data only, and high computational complexity. To overcome the above weaknesses of the state-of-the-art (SOTA) methods, our work made the following key contributions:
\begin{itemize}
\item We propose a novel framework for detecting poisoning attacks in FL, which employs a reference model (used to check stitching connectivity) based on a public dataset and an auditor model to detect malicious updates. It can reach the lowest possible computational complexity $\mathcal{O}(K)$ where $K$ is the number of clients.

\item We implemented a detector based on the proposed framework and using a one-class support vector machine (OC-SVM) and made it open source [URL to be released after acceptance].

\item We evaluated our detector's performance against SOTA poisoning attacks for two typical applications of FL: ECG classification and human activity recognition (HAR). Our experimental results showed that our detector can indeed overcome all the above-mentioned weaknesses.
\end{itemize}

The rest of the paper is organized as follows. The next section introduces the background and related work. Section~\ref{sec:proposed_framework} presents the proposed framework. Sections~\ref{sec:performance} and \ref{sec:comparision} cover the performance evaluation and comparison of the proposed framework, respectively. Section~\ref{sec:further_discussions} presents some further discussions and discusses the limitations of our work. Finally, the last section concludes the paper.

\section{Background and Related Work}
\label{sec:background}

\subsection{Federated Learning}

Federated learning (FL)~\cite{mcmahan2016federated} is a concept of distributed machine learning in which different edge devices or silos (hospitals, companies, etc.) collaborate to train a joint model known as a global model. The global model is trained without requiring participating edge devices to share their local raw data with other participating nodes or a centralized server. Instead, each edge device trains its local model using its local data and shares only the trained parameters with a central server in centralized FL or with all other nodes in decentralized FL. The aggregation of local models is done according to a given aggregation algorithm to compute parameters of the global model~\cite{zhang2021}, among which FedAvg~\cite{mcmahan2017} is the most commonly used algorithm~\cite{sun2021}. Mathematically, FedAvg is given by the following equation:
\begin{equation}
\textit{AW}=\frac{n_k}{n_K}\sum_{k=1}^KW_k,
\end{equation}
where \textit{AW} is the global aggregated weights, $n_k$ and $n_K$ are the number of samples of an individual edge and the total number of samples of all the edge devices taking part in the global round, respectively. $W_k^t$ are the locally trained weights of the $k$-th edged device/client, and $K$ is the number of total edged devices/clients taking part in the global round.

\subsection{Stitching Connectivity}

Stitching connectivity~\cite{bansal2021} is a method to measure the similarity of internal representations of different models trained using different but similar data. Consider two models \textit{A} and \textit{B}, which have the same architecture. For \textit{A} and \textit{B} to be stitched connected, they can be stitched at all the layers to each other. In other words, two models, let's say \textit{A} and \textit{B} with identical architecture but trained using stochastic gradient descent(or its variants) using independent random seeds and independent training sets taken from the same distribution. Then the two trained models are stitched and connected for natural architectures and data distributions. Hence, we expect the models trained on similar but different training sets of the same distribution will behave similarly. Based on this property, we adopt Proposition~\ref{prop:noisy_data_training}.

\begin{prop}
\label{prop:noisy_data_training}
When a model is trained on noisy data (malicious/poisoned), the first half of the layers are similar to a model trained on good-quality data (benign).
\end{prop}

\subsection{Memorization in deep networks}

Deep neural networks are capable of memorizing the training data in a fashion such that they prioritize learning simple patterns first~\cite{arpit2017} using the lower-level layers in the model, while the higher-level layers tend to learn more specific data characteristics. Furthermore, when a model is trained on noisy data, the first half of the layers are similar to a model trained on good-quality data~\cite{bansal2021}. Based on this, we adopt Proposition~\ref{prop:activations}.

\begin{prop}
\label{prop:activations}
Different models with the same architecture but random initial seeds, trained on different training sets of a similar distribution, have similar internal representations and thereafter similar activations for a given test input sample.
\end{prop}

\subsection{Byzantine attacks}
\label{sec:background_attacks}

Byzantine attacks are a type of attack where a trusted device or a set of devices turn rogue and try to compromise the overall system. Such attacks can significantly reduce the performance of the global model in federated learning~\cite{Hu2022byzantine}. Researchers have shown that in the case of some aggregation algorithms, even the presence of a single malicious node can significantly reduce the performance of the global model~\cite{fang2020, blanchard2017, li2020}. Byzantine attacks, such as poisoning attacks~\cite{tolpegin2020, mammen2021}, can substantially reduce the performance (classification accuracy, precision, and recall) of FedAvg, even in the presence of a very small percentage of adversarial participants in the network. Such attacks can be classified as targeted attacks that negatively impact only one or more target (but not all) classes under attack, and untargeted attacks that impact all the classes negatively. Furthermore, poisoning attacks are mainly classified into two main categories: data poisoning~\cite{tolpegin2020} and model poisoning~\cite{fang2020} attacks depending on the phase where the attacks are executed. If the attacker manipulates the training data then this is called data poisoning attacks and if the attacker manipulates the trained model's parameters then such attacks are called model poisoning attacks. Further details of each type of poisoning are given as follows:

\textbf{Data poisoning attacks:}

Data poisoning attacks~\cite{tolpegin2020, Hu2022byzantine,sun2021data} are those attacks in which the attacker manipulates the training data directly according to a given strategy and then trains the model using the manipulated dataset. In this study, we consider the following four types of SOTA data poisoning attacks:
\begin{enumerate}
\item \textbf{Random label flipping poisoning attacks:} In such attacks the attacker flips the true labels of the training instance randomly.

\item \textbf{Random label and feature poisoning attacks:} In such attacks, in addition to flipping the label randomly, the attacker adds noise to the input features of the training instances.

\item \textbf{Label swapping poisoning attacks:} In such attacks the attacker swaps the labels of selected samples of a given class with those of another class.

\item \textbf{Feature poisoning attacks:} In such attacks the attacker adds noise (enough to manipulate the global model) to the features of the training data.
\end{enumerate}

\textbf{Model poisoning attacks:}

In the model poisoning attacks~\cite{fang2020} the attacker trains the model using legitimate datasets and then manipulates the learned parameters before sending it to the global server. In this study, we consider the following four types of SOTA model poisoning attacks:
\begin{enumerate}
\item \textbf{Sign flipping attacks:} In such attacks the attacker trains the model using the legitimate data and then flips the sign of trained parameters and enlarges their magnitude.

\item \textbf{Same value attacks:} In such attacks the attacker sets the parameter values as \textbf{C}, where \textbf{C} corresponds to a vector whose elements have an identical value C, which is a constant set to a value such as 100, 200, 300, etc.

\item \textbf{Additive Gaussian noise attacks:} In such attacks the attacker trains the model as expected with legitimate data but adds Gaussian noise before sharing the updates with the global server.

\item \textbf{Gradient ascent attacks:} In such attacks the attacker trains the models using a gradient ascent instead of a gradient descent optimizer.
\end{enumerate}

\subsection{Existing defense mechanisms}

To address the above-mentioned attacks, many defense methods have been proposed. Hu et al.~\cite{Hu2022byzantine} surveyed the state-of-the-art defensive methods against byzantine attacks in FL. They show that byzantine attacks by malicious clients can significantly reduce the accuracy of the global model. Moreover, their results show that existing defense solutions cannot fully protect FL against such attacks. In \cite{Xia2023} Xia et al.\ summarized poisoning attacks and defense strategies according to their methods and targets. In this section, we discuss some of the existing SOTA defense mechanisms below.

\subsubsection{Distance-Based Mechanisms}

Such mechanisms detect malicious updates by calculating the distance between updates. Updates with a larger distance from others are discarded from global aggregation. For example, Blanchard et al.~\cite{blanchard2017} proposed Krum, where the central server selects updates with minimum distance from the neighbors. Similarly, Xia et al.~\cite{xia2019faba} proposed a method to discard the shared parameters with a large distance from the mean of shared parameters. Bo et al.~\cite{Bo2022FedInv} proposed FedInv, which inverses the updates from each client to generate a dummy dataset. The server then calculates Wasserstein distances for each client and removes the update(s) with exceptional Wasserstein distances from others. Gupta et al.~\cite{gupta2022long} proposed a method called MUD-HoG, a method to address poisoning attacks in federated learning using a long-short history of gradients of clients. Nevertheless, for such methods to work the number of attackers is needed to be known in advance or needs to be less than a certain percentage of total clients in the network.

\subsubsection{Performance-Based Mechanisms}

Methods in this category evaluate updates based on a clean dataset that is contained by the server. Updates that underperform are either assigned low weights or removed from the global aggregation. For example, Li et al.~\cite{li2019} proposed using a pre-trained autoencoder to evaluate the performance of an update. The performance of malicious updates will be lower as compared to that of benign updates. Nevertheless, training an autoencoder needs sufficient benign model updates which are hard to get. Xie et al.~\cite{xie2019} proposed Zeno, which requires a small dataset on the server side. It computes the score for each update using the validation dataset server-side. A higher score implies a higher probability of the respective update being benign and vice-versa. However, Zeno requires knowledge about the number of attackers in advance to work properly.

\subsubsection{Statistical Mechanisms}

A method in this category uses statistical features of the shared gradients or updates. Commonly used features are mean or median. Such features can circumvent benign updates and help to achieve a robust and probably more benign gradient. For example, a coordinate-wise median and a coordinate-wise median-based solution are provided by Yin et al.~\cite{yin2018}, which aggregates the parameters of local models independently, i.e., for the $i$-th model parameter the global server sorts the $i$-th parameter of the $m$ other local models. Removes the largest and smallest parameters and computes the mean of the remaining parameters as the $i$-th parameter of the global model. Similarly, El~Mhamdi et al.~\cite{guerraoui2018} proposed to combine Krum and a variant of trimmed mean~\cite{yin2018}. However, these approaches are vulnerable to poisoning attacks even using robust aggregation~\cite{liu2022threats}. Additionally, for such methods to work, the number of attackers should be limited by an upper bound (e.g., 50\%) or should be known prior.

\subsubsection{Target Optimization-Based Mechanisms}

Target optimization-based methods optimize an objective function to improve the robustness of the global model. For example, Li et al.~\cite{li2019rsa} RSA, which regularizes the objective in such a way that it forces each local model in federated learning to be close to the global model. However, such methods only address data poisoning attacks and fail to eliminate model poisoning attacks.

\section{Proposed Framework}
\label{sec:proposed_framework}

\subsection{Threat Model}

\textbf{Attacker's goal:} Similar to many other studies~\cite{biggio2012, Chen2021, shafahi2018}, we consider an attacker whose goal is to manipulate the global model in such a way that it has low performance (i.e., high error rates) and/or misbehaves in a particular way. Such attacks make the global model underperform. For example, an attacker can attack competitor FL systems. We consider both targeted and untargeted~\cite{suciu2018} attacks, as discussed previously.

\textbf{Assumptions:} We consider the following assumptions about the threat model:
\begin{enumerate}
\item There are one or more malicious edge devices that try to launch a model poisoning attack against the global model, possibly in a collaborative manner (i.e., launching a colluding attack).

\item All the attackers follow the FL algorithm, i.e., they train their local model using their local data and share the parameters with the global model.
    
\item All the attackers have two strategies: manipulating their local training data and training the local model using the manipulated data, and manipulating the model parameters after training it on benign data.
    
\item All the attackers have the full knowledge about the aggregation rules, the global model architecture, the auditor model (see Section~\ref{subsec:overview} for more details about the auditor model proposed in our method), and the detection results, i.e., the whole detection framework is a white box and can be used as an oracle to adapt the attackers' strategy.
    
\item The global server can trust a third party or an isolated component, which maintains a public dataset that has data representing all classes of the underlying application for training the auditor model, and the attackers cannot poison this public dataset or the auditor model.
\end{enumerate}

Note that the final assumption may look very strong, but the existence of a public dataset is common in many fields (e.g., a public health dataset maintained by the scientific community).

\subsection{Overview}
\label{subsec:overview}

\begin{figure*}
\centering
\includegraphics[width=\linewidth]{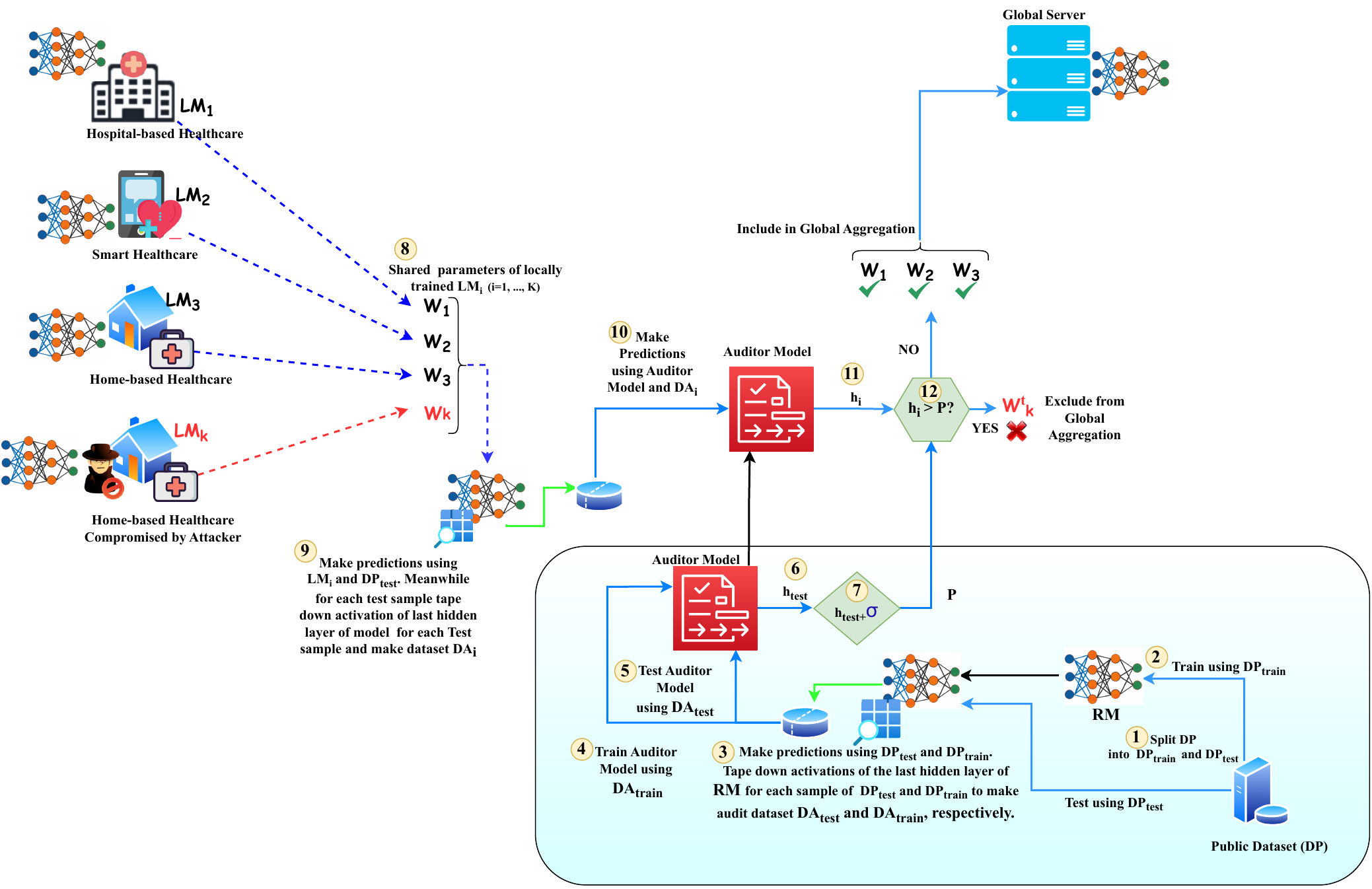}
\caption{Overview of the proposed framework}
\label{fig:overview}
\end{figure*}

In this subsection, we describe the proposed framework. An overview of the proposed method has been shown in Figure~\ref{fig:overview}. Let us assume $K$ edged devices (hospitals, organizations, etc.) collaborate to train a joint global model \textbf{GM}. An edge $\textbf{E}_k$ trains a local model $\textbf{LM}_k$ using its local data $\textbf{D}_k$, where $k={1, 2, \dots, K}$. The global server \textbf{GS} is responsible for receiving the updates from edge devices and aggregation. We assume that a trusted third party or \textbf{GS} (We consider a trusted third party to be a different entity from \textbf{GS}) also has an open-source dataset which is called public data and we represent it as \textbf{DP}. \textbf{DP} is supposed to be a representative dataset of all the classes in a classification problem, i.e., it has samples from each candidate class. The training for a global round is given as follows.
\begin{enumerate}
\item Trusted third party creates an audit model $\textbf{AM}$ and a reference model $\textbf{RM}$. Where $\textbf{RM}$ has the same architecture as global model \textbf{GM}.

\item Trusted third party splits \textbf{DP} into train $\textbf{DP}_\text{train}$ and test $\textbf{DP}_\text{test}$ datasets and trains the $\textbf{RM}$ using $\textbf{DP}_\text{train}$.

\item After training, trusted third party makes predictions with $\textbf{RM}$ using $\textbf{DP}_\text{train}$ and  $\textbf{DP}_\text{test}$. During the predictions for each dataset, a trusted third party taps the activations of the last hidden layer for each input sample using Algorithm~\ref{algo:algorithm} to create a dataset $\textbf{DA}_\text{train}$, and $\textbf{DA}_\text{test}$ for each $\textbf{DP}_\text{train}$ and  $\textbf{DP}_\text{test}$, respectively. 

\item Trusted third-party trains the audit model (a one-class classifier) using $\textbf{DA}_\text{train}$. Here, we treat all the samples of $\textbf{DA}_\text{train}$ as a single class. Trusted third party sends the trained $\textbf{AM}$, $\textbf{RM}$, $\textbf{DP}_\text{test}$ and a value $P$ to $\textbf{GS}$. We define $P$ later in the section. 

\item Each $\textbf{E}_i$ trains $\textbf{LM}_i$ using $\textbf{D}_i$.

\item Each $\textbf{E}_i$ sends updates $W_i$ of trained $\textbf{LM}_i$ to \textbf{GS}.

\item \textbf{GS} sets $W_i$ as the parameters of $\textbf{RM}$ and makes predictions using $\textbf{DP}_\text{test}$. During the predictions, \textbf{GS} taps the activations of $\textbf{RM}$ for each input sample using Algorithm~\ref{algo:algorithm} to create a dataset $\textbf{DA}_i$.

\item \textbf{GS} makes predictions using \textbf{AM} and $\textbf{DA}_i$ as input data. For every input sample $X \in \textbf{DA}_i$, $\textbf{AM}$ outputs $y_i\in\{1,-1\}$ and creates a set $Y=\{y_1, y_2, \dots, y_z\}$, where $z$ is the total number of samples in $\textbf{DA}_i$.

\item \textbf{GS} computes poisoned rate $h_i=\frac{o\times100}{z}$, where $o$ is the total number of $-1's$ in $Y$.

\item \textbf{GS} includes $W_i$ in global aggregation if $h_i\le P$, otherwise discards $W_i$. Here, $P=h_\text{test}+\alpha\sigma$, and it is the percentage of poison that we want to tolerate. $\sigma$ is called deviation tolerance and it is given as $\sigma=|h_\text{test} - h_\text{train}|$. $h_\text{test}$ and $ h_\text{train}$ are calculated using $\textbf{DP}_\text{train}$ and  $\textbf{DP}_\text{test}$, respectively. $\alpha$ is a parameter to make the threshold flexible. In our experiments, we set $\alpha=1$.
\end{enumerate}

\begin{algorithm}[!ht]
\caption{Creation of Audit Dataset(s) $\textbf{DA}_i$}
\label{algo:algorithm}
\DontPrintSemicolon

\KwInput{A dataset $\textbf{D}_i$ and a trained model \textbf{M}} \KwOutput{a new dataset $\textbf{DA}_i$}
\For { $X \in \textbf{D}_i$}
{Transform $X$ into batch size\\
 Make Prediction using $X$ as a test sample in \textbf{M}\\
 Get activation maps $A_{l,w}^k$ of the last convolutional layer. where $k$ is the number of activation maps with length $l$ and width $w$ each.\\
 Get probability score $y^{c}$, where $c$ is the class label of $X$.\\ 
 Reshape $A^k$ as an array of $1\times j$, where $j=l\times w\times k$.\\
 Reshape $X$ as an array of $1\times l$, where $l$ is the product of the height and width of $X$\\
 Compute $s= X \mathbin\Vert A^k \mathbin\Vert y^c$\\
 Append $s$ in $\textbf{DA}_i$ as a new data sample}
 Output $\textbf{DA}_i$

\end{algorithm}

The proposed framework works based on two propositions that we discussed previously.

Based on Proposition~\ref{prop:activations}, we expect that the models trained on different datasets of similar distributions will behave similarly. Since the higher-level layers learn the information specific to the data of the model, we take the activation of the last hidden layer (in the case of CNN's last convolutional layer) to capture more information (features) related to the training data. We only consider the activations of the last as it provides sufficient information about the underlying training data to detect poisoning attacks. Activations from other higher-level layers can also be incorporated for this purpose, nevertheless, this increases the computational costs with no significant improvement in the results for detecting poisoning attacks. Hence, we only consider the activations of the last convolutional hidden layer. We capture this property by training another model (\textbf{AM}) to learn the behavior of a model trained on the benign dataset (\textbf{RM} in our case). Hence, models behaving similarly to the model trained on the benign datasets are probably free from poisoning attacks, and at least they do not degrade the performance of the global model.

\section{Performance Evaluation}
\label{sec:performance}

We evaluated the proposed framework using two healthcare applications, i.e., ECG classification and HAR.

\subsection{Experimental Setup}

\textbf{Datasets:} For ECG classification, we use the widely known MIT-BIT arrhythmia dataset~\cite{moody2001}. The dataset contains 48-half-hour two-channel ECG recordings. These recordings were obtained from 47 subjects. The dataset contains 109,446 samples, sampled at a frequency of 125 Hz. Further, the dataset contains five classes of ECG: non-ecotic beats (normal beats), supraventricular ectopic beats, ventricular ectopic beats, fusion beats, and unknown beats.

For HAR, we used the dataset in~\cite{raza2021, kwapisz2011}. The dataset contains time-series data related to 14 different human activities (standing, sitting, walking, jogging, going up-stairs, going down-stairs, eating, writing, using a laptop, washing face, washing hands, swiping, vacuuming, dusting, and brushing teeth) collected using sensors such as accelerometers, magnetometers, and gyroscopes.

\textbf{Classifiers:} We developed a convolution neural networks (CNN) based classifier for each application. The developed classifiers do not achieve the optimum classification for the considered datasets. This is because our objective is to show that our proposed framework can detect anomalous (poisoning attacks), not to achieve the best performance in terms of classification. For ECG classification, we developed a five-class classifier, and the aim of FL here is to learn a global five-class classifier. for HAR, we developed a fourteen-class classifier, and the aim of FL here is to learn a global fourteen-class classifier.

\textbf{Federated Setting:} To simulate the federated setting, we simulated a network with three edge devices (two benign ones and one attacker) where the three edge devices were implemented using Tensorflow 2.11 and Python 3 in a Dell latitude laptop with $12^{th}$ Gen Intel® Core™ i7-1265U processor and 16 MB DDR4 RAM and a global server using Tensorflow 2.11 and Python 3 in a Dell workstation with 32 GB RAM and an Intel® Core™ i-6700HQ CPU. 

\subsection{Performance Evaluation}

To evaluate the performance of the proposed framework, we first tested the proposed framework to check its ability to differentiate samples in $\textbf{DA}_i$ of benign edges from samples in $\textbf{DA}_i$ of malicious edges. We trained a \textbf{RM} and an \textbf{AM} using a public dataset (which is excluded from the training data of the edge devices and the test data). We distributed each candidate dataset (ECG and HAR) equally (in the case of the HAR dataset, benign nodes contain slightly more data) but randomly among the benign and malicious nodes. For each dataset, we did the following. We created $\textbf{DA}_i$ using each benign edge and labeled each sample as 1 (for benign). Similarly, for the malicious edge, we created $\textbf{DA}_i$ using each type of model and data poisoning attack discussed in Section~\ref{sec:background_attacks} and selected random samples from each attack's $\textbf{DA}_i$ to make a new $\textbf{DA}_i$ and labeled each sample as -1. We combined the two $\textbf{DA}_i$'s that we created. Then, this lastly created dataset was used to test the \textbf{AM}. Note that here we just show the ability of the proposed framework to classify benign and malicious samples not the detection of updates. We will show the detection of updates later. Table~\ref{tab:calssification_accuracy_ecg} shows the classification accuracy of the proposed framework to differentiate samples generated using shared parameters of benign edged devices and the samples generated using shared parameters of malicious edge devices for ECG classification. Similarly, Table~\ref{tab:calssification_accuracy_har} shows the classification accuracy of the proposed framework to differentiate samples generated using shared parameters of benign edged devices and the samples generated using shared parameters of malicious edge devices for HAR. For both types of applications, it can be seen that the proposed framework can differentiate samples of benign and malicious edged devices very well, with an overall accuracy of 94\% and 99\% for HAR and ECG classification, respectively. As mentioned previously, here we calculate the overall accuracy of \textbf{AM} to classify benign and malicious samples which is 94\% and 99\%  for HAR and ECG respectively. It should be noted that the poison attack detection is still 100\% as updates with  94\% and 99\% poisoned samples will have a poison rate of around 94\% and 99\%, respectively. Hence, updates with 94\% and 99\% poison rates will be classified as malicious and will be removed from global aggregation. 

\begin{table}[!ht]
\centering
\caption{Classification accuracy of \textbf{AM} for ECG classification}
\label{tab:calssification_accuracy_ecg}
\begin{tabular}{ccccc}
\toprule
\textit{Class} & \textit{Precision} & \textit{Recall} & \textit{F1-Score} & \textit{\#(Samples)}\\
\midrule
Benign  & 100 & 97 & 98 & 9,000\\
Malicious & 97 & 100 & 99 & 9,000\\
Accuracy & & & 99 & 18,000\\
Micro average & 99 & 99 & 99 & 18,000\\
Weighted average & 99 & 99 & 99 & 18,000\\
\bottomrule
\end{tabular}
\end{table}

\begin{table}[!ht]
\centering
\caption{Classification accuracy of \textbf{AM} for HAR}
\label{tab:calssification_accuracy_har}
\begin{tabular}{ccccc}
\toprule
\textit{Class} & \textit{Precision} & \textit{Recall} & \textit{F1-Score} & \textit{Support}\\
\midrule
Benign (1) & 96 & 95 & 95 & 22,280\\
Malicious (-1) & 90 & 91 & 91 & 11,140\\
Accuracy & & & 94 & 33,420\\
Micro average & 93 & 93 & 93 & 33,420\\
Weighted average & 94 & 94 & 94 & 33,420\\
\bottomrule
\end{tabular}
\end{table}

Now, we show the ability of the proposed framework to detect malicious updates and remove them from the global aggregation. We simulated the proposed framework as follows. First, the trusted third party prepares the \textbf{RM} and the \textbf{AM} and trains the \textbf{RM} and the \textbf{AM} using the public data and sends it to \textbf{GS}. Then the benign devices and the attacker send the updates to the \textbf{GS}. The attacker launches a given attack at each global round. For example, for round $r$ it uses a random label flipping attack and for $r+1$ the attacker launches a feature poisoning attack, and so on. We tested the performance of our proposed framework under all the different types of model and data poisoning attacks listed in Section~\ref{sec:background_attacks}. Moreover, we distributed the data among the edged devices randomly but using non-independent and identically distributed unbalanced data distribution, where each edge has 40-50\% fewer data for a given class. For example, edge 1 has 40\% fewer samples of non-ecotic beats class for the ECG dataset as compared to other edge devices, and edge 2 has 50\% fewer samples of supraventricular ectopic beats class compared to other edge devices. In comparison, edge 3 (attacker) has an equal number of samples for each class. This is done to give more attacking power to the edge3. For example, if the attacker has an equal number of samples of each class it can launch a random label-flipping attack which will eventually affect all the classes and will affect the global model significantly. We follow a similar distribution for the HAR dataset. 

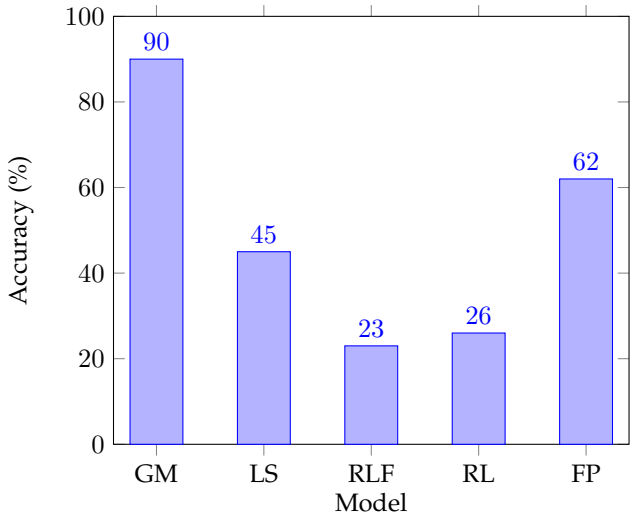
\begin{figure}
\centering
\begin{tikzpicture}
\begin{axis}
[
ybar,
ymin=0,
ymax=100,
xlabel=Model,
ylabel={Accuracy (\%)},
bar width=20pt,
symbolic x coords={GM, LS, RLF, RL, FP},
xticklabel style={text height=1.5ex}, 
xtick=data,
nodes near coords,
nodes near coords align={vertical}]
\addplot coordinates {
(GM,    90)     
(LS,    45)     
(RLF,   23)    
(RL,    26)  
(FP,    62)};
\end{axis}
\end{tikzpicture}
\caption{Performance of the global model with and without proposed framework under four different data poisoning attacks on HAR.}
\label{fig:har_data_poisoning}
\end{figure}

First, we present the comparison of the accuracy of the global model after one global round with and without our proposed framework under four different data poisoning attacks. Note that our global model achieves optimum accuracy after one global round, hence we launched the attack in each global round. Otherwise, the attacks can be launched at any given global round. Figure~\ref{fig:ecg_data_poisoning} presents a comparison of the accuracy of the global model under different data poisoning attacks and with and without the proposed framework for ECG classification. Here, \textit{GM} is a global model with our proposed framework under different data poisoning attacks, \textit{LS} represents the global model under a label swapping attack without our proposed framework, \textit{RLF} represents the global model under a random label and feature poisoning attack without our proposed framework, \textit{RL} represents the global model under a random label attack without our proposed framework, and \textit{FP} represents the global model under a feature poisoning attack without our proposed framework, all in a federated setting. It can be seen that the performance of the global model deteriorates significantly under the data poisoning attacks. It can also be seen that the performance of the global model was not affected by the poisoning attacks when we applied our proposed framework that could detect and eliminate almost all poisoned updates from the malicious edge before global aggregation. As mentioned previously, even if the accuracy of \textbf{AM} for benign and malicious sample detection is not 100\%, it can accurately detect all the malicious updates because the poisoning rate is high. Hence, its malicious updates detection is still 100\%. Similarly, Figure~\ref{fig:har_data_poisoning} presents the performance of the global model compared under four different data poisoning attacks with and without adopting the proposed framework for HAR. Since all the malicious updates are detected and removed before aggregation, the accuracy of the global model remains almost similar for a given dataset.

\begin{figure}
\centering
\begin{tikzpicture}
\begin{axis}
[
ybar,
ymin=0,
ymax=100,
xlabel=Model,
ylabel={Accuracy (\%)},
bar width=20pt,
symbolic x coords={GM, LS, RLF, RL, FP},
xticklabel style={text height=1.5ex}, 
xtick=data,
nodes near coords,
nodes near coords align={vertical}]
\addplot coordinates {
(GM,    87)     
(LS,    43)     
(RLF,   37)    
(RL,    26)  
(FP,    76)};
\end{axis}
\end{tikzpicture}
\caption{Performance of the global model with and without our proposed framework under four different data poisoning attacks on ECG classification. Here GM is Global Model, LS is Label swapping attack, RLF is random Label Flipping, RL is random Label and FP is Feature poisoning.}
\label{fig:ecg_data_poisoning}
\end{figure}

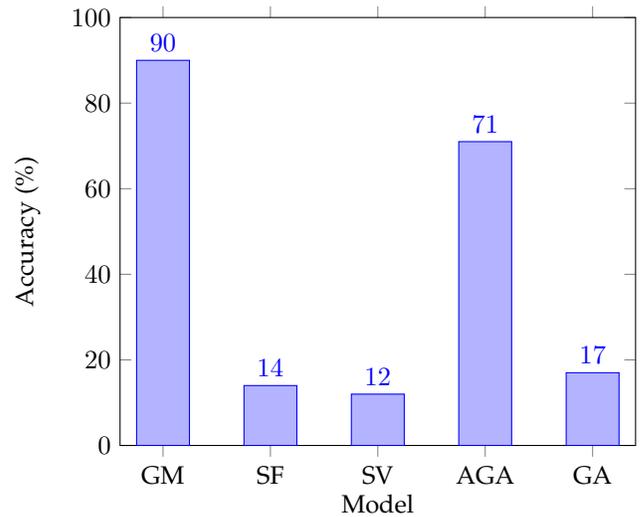
\begin{figure}
\centering
\begin{tikzpicture}
\begin{axis}
[
ybar,
ymin=0,
ymax=100,
xlabel=Model,
ylabel={Accuracy (\%)},
bar width=20pt,
symbolic x coords={GM, SF, SV, AGA, GA},
xticklabel style={text height=1.5ex}, 
xtick=data,
nodes near coords,
nodes near coords align={vertical}]
\addplot coordinates {  
(GM,         90)
(SF,         14)   
(SV,         12)   
(AGA,        71)    
(GA,         17)};
\end{axis}
\end{tikzpicture}
\caption{Performance of the global model with and without proposed framework under four different model poisoning attacks on HAR.}
\label{fig:har_model_poisoning}
\end{figure}

Moreover, Figure~\ref{fig:ecg_model_poisoning} compares the global model's accuracy under four different model poisoning attacks with and without our proposed framework for ECG classification. Here, \textit{GM} shows the performance of the global model with our proposed framework, \textit{SF} shows the performance of the global model under a sign flip attack without our proposed framework, \textit{SV} shows the performance of the global model under a same value attack without our proposed framework, \textit{AGA} shows the performance of the global model under a performance additive Gaussian noise attack without a proposed framework, and \textit{GA} shows the performance of the global model under a gradient ascent attack without our proposed framework, all in a federated setting. Similarly, Figure~\ref{fig:har_model_poisoning} compares the global model's accuracy under the four different model poisoning attacks with and without our proposed framework for HAR.

\begin{figure}
\centering
\begin{tikzpicture}
\begin{axis}
[
ybar,
ymin=0,
ymax=100,
xlabel=Model,
ylabel={Accuracy (\%)},
bar width=20pt,
symbolic x coords={GM, SF, SV, AGA, GA},
xticklabel style={text height=1.5ex}, 
xtick=data,
nodes near coords,
nodes near coords align={vertical}]
\addplot coordinates {  
(GM,         88)
(SF,         18)   
(SV,         07)   
(AGA,        67)    
(GA,         21)};
\end{axis}
\end{tikzpicture}
\caption{Performance of the global model with and without proposed framework under four different model poisoning attacks on ECG classification. Here GM is Global Model,SF Sign
Flip attack, SV is Same Value attack, AGA Additive
Gaussian noise Attack, GA is Gradient Ascent attack  }
\label{fig:ecg_model_poisoning}
\end{figure}

 Table~\ref{tab:detection_data_poisoning_har} shows the poisoning rate identified by the proposed framework for each edge in the federated setting for HAR. Here Edge 1 and Edge 2 are benign devices and Edge 3 is the malicious device that launches different data poisoning attacks in each global round, whereas the $P$ value in the table shows the threshold, which is calculated using $h_\text{test}$ and $\sigma$. Any updates ($W_i$) from an edge device for which its corresponding $DA_i$ has a value of $h_i$ greater than $P$ will be marked as malicious or poisoned and removed from global aggregation. It can be seen that in Table~\ref{tab:detection_data_poisoning_har}, $h_1$ and $h_2$ for Edge 1 and Edge 2, respectively, have a value smaller than $P$. Hence, the updates $W_1$ and $W_2$ will be included in global aggregation, while $W_3$ will be excluded from global aggregation as its corresponding $h_3$ is greater than $P$. It can be seen that our proposed framework can not only detect the attacks but also provide insights into the percentage (a general idea about possible poison percentage) of poisoned data being used to train a malicious local model. For example, for LS, we swapped the labels of some classes (for ECG classification two classes and HAR 10 classes) while keeping the rest of the labels in their original form; thereafter, it gives a value of 47.7 for ECG and 81.2 for HAR. Similarly, Tables~\ref{tab:detection_data_poisoning_ecg}, \ref{tab:detection_data_poisoning} and \ref{tab:detection_model_poisoning_har} present the performance of our proposed framework to detect data poisoning attacks on ECG classification, model poisoning attacks on ECG classification, and model poisoning attacks on HAR, respectively. Hence, these results show that the proposed framework can identify malicious updates which are then removed from the global aggregation. 

\begin{table*}[!ht]
\centering
\caption{Detection of data poisoning attacks on HAR}
\label{tab:detection_data_poisoning_har}
\begin{tabular}{ccccc}
\toprule
\textit{Attack Type} & $P$ ($h_{\text{test}}=20.0, \sigma=10.0$)  & \textit{Edge 1 ($h_1$)} & \textit{Edge 2 ($h_2)$} & \textit{Edge 3 (Malicious) $(h_3)$}\\
\midrule
RLF & 30.0 & 23.3 & 11.5 & 94.5\\
RL & 30.0 & 23.5 & 10.0 & 100\\
LS & 30.0 & 21.1 & 19.7 & 81.2\\
FP & 30.0 & 18.3 & 16.1 & 91.0\\
\bottomrule
\end{tabular}
\end{table*}

\begin{table*}[!ht]
\centering
\caption{Detection of data poisoning attacks on ECG classification}
\label{tab:detection_data_poisoning_ecg}
\begin{tabular}{ccccc}
\toprule
\textit{Attack Type} & $P$ ($h_\text{test}=15.0, \sigma=10.0$)  & \textit{Edge 1 ($h_1$)} & \textit{Edge 2 ($h_2)$} & \textit{Edge 3 (Malicious) ($h_3)$}\\
\midrule
RLF & 25.0 & 10.2 & 14.5 & 100\\
RL & 25.0 & 14.5 & 19.0 & 100\\
LS & 25.0 & 12.1 & 21.3 & 47.7\\
FP & 25.0 & 13.3 & 22.1 & 100\\
\bottomrule
\end{tabular}
\end{table*}

\begin{table*}[!ht]
\centering
\caption{Detection of model poisoning attacks on HAR classification}
\label{tab:detection_model_poisoning_har}
\begin{tabular}{ccccc}
\toprule
\textit{Attack Type} & $P$ ($h_\text{test}=20.0, \sigma=10.0$)  & \textit{Edge 1 ($h_1$)} & \textit{Edge 2 ($h_2)$}& \textit{Edge 3 (Malicious) $(h_3)$}\\
\midrule
SF & 30.0 & 20.3 & 19.5 & 100\\
SV & 30.0 & 20.2 & 20.1 & 100\\
AGA & 30.0 & 20.1 & 19.7 & 90.1\\
GA & 30.0 & 20.3 & 20.1 & 100\\
\bottomrule
\end{tabular}
\end{table*}

\begin{table*}[!ht]
\centering
\caption{Detection of model poisoning attacks on ECG classification}
\label{tab:detection_data_poisoning}
\begin{tabular}{ccccc}
\toprule
\textit{Attack Type} & $P$ ($h_\text{test}=15.0, \sigma=10.0$)  & \textit{Edge 1 ($h_1$)} & \textit{Edge 2 ($h_2)$}& \textit{Edge 3 (Malicious) $(h_3)$}\\
\midrule
SF & 25.0 & 10.0 & 11.7 & 100\\
SV & 25.0 & 10.2 & 10.1 & 100\\
AGA & 25.0 & 9.9 & 9.7 & 96.1\\
GA & 25.0 & 10.1 & 10.4 & 100\\
\bottomrule
\end{tabular}
\end{table*}

\section{Comparison}
\label{sec:comparision}

Table~\ref{tab:caomparision} shows a comparison of our proposed framework with some of the SOTA methods for detecting poisoning attacks in FL~\cite{xia2019faba, cao2019, guerraoui2018, blanchard2017, fung2018, li2019, xie2019, cao2019dis, munoz2019byzantine, li2019rsa, pillutla2019robust, Bo2022FedInv}, where the \textit{Category} column presents the type of detection mechanism, attack type presents a type of attack, the \textit{Model Accuracy} column presents the accuracy of global model under attack compared with the accuracy of the global model without any attack, the \textit{Data Distribution} column presents the type of data distribution among clients: independent identically distributed (IID), non identically distributed (Non-IID), $d$ represents model size (depth) of each client and $K$ is the total number of edge devices/clients. It can be seen that our proposed framework provides more desirable features than all other SOTA methods. Our proposed framework does not need knowledge about the number of attackers. It can detect both types of attacks, i.e., model and data poisoning attacks. Moreover, its time complexity is the least compared to others. Additionally, the proposed method can be used in both IID and non-IID data distribution with high accuracy.

\begin{table*}[!ht]
\centering
\caption{Comparison with SOTA methods.}
\label{tab:caomparision}
\begin{tabular}{cM{0.15\linewidth}cM{0.15\linewidth}cccc}
\toprule
\textit{Scheme (Year)} & \textit{Category}  & \textit{Attack Type} & \textit{Total Number of Attackers} & \textit{Model Accuracy} & \textit{Data Distribution} & \textit{Time Complexity}\\
\midrule 
\cite{xia2019faba} (2019)& Distance based & Data/Model & Less than 50\% & Medium & IID&$\mathcal{O}({K^{2}d})$\\
\cite{cao2019} (2019)& Distance based & Data & Less than 30\% & Medium & IID& $\mathcal{O}({K^{2}d})$\\
\cite{guerraoui2018} (2018)& Statistic based & Data/ Model & Less than 50\% & High & IID/Non-IID&$\mathcal{O}({K^{2}d})$\\
\cite{blanchard2017} (2017)& Distance based & Data/Model & Less than 50\% & Medium & IID&$\mathcal{O}({K^{2}d})$\\
\cite{fung2018} (2018) & Distance-based & Data/Model& No limitation& High/Medium & IID/Non-IID&$\mathcal{O}({K^{2}d})$\\
\cite{li2019} (2019)& Performance-based & Data/Model& less than 50\% & High & IID/Non-IID&$\mathcal{O}({Kd})$\\
\cite{xie2019} (2019)& Performance-based & Data/Model& one honest user (At-least) & High & IID/Non-IID&$\mathcal{O}({Kd})$\\
\cite{cao2019dis} (2019) & Performance-based & Data/Model& No limitation & High & IID&$\mathcal{O}({Kd})$\\
\cite{munoz2019byzantine} (2019) & Statistic based & Data/Model& less than 50\% & High & IID&$\mathcal{O}({K^{2}d})$\\
\cite{li2019rsa} (2019)& Target optimization based & Data/Model& No limitation & High & IID&$\mathcal{O}({Kd})$\\
\cite{pillutla2019robust} (2019) & statistic based
 & Data/Model& less than 50\% & Medium & IID&$\mathcal{O}({Kd})$\\
\cite{Bo2022FedInv} (2022) & Distance based & Data & less than 50\% & High & IID/Non-IID&$\mathcal{O}({Kd})$\\
\cite{gupta2022long} (2022) & Distance based & Data/Model & less than 50\% & High & IID/Non-IID&$\mathcal{O}({Kd})$\\
\cite{xutdfl2022} (2022)& Distance based & Data/Model & less than 50\% or known number of attackers & High & IID/Non-IID&$\mathcal{O}({Kd})$\\
\cite{jeong2022fedcc} (2022)& Distance based & Model & less than 50\%& High & IID/Non-IID&$\mathcal{O}({Kd})$\\
\midrule
Proposed & Machine learning based & Data/Model & No limitation & High & IID/Non-IID&$\mathcal{O}(K)$\\
\bottomrule
\end{tabular}
\end{table*}

\section{Further Discussions}
\label{sec:further_discussions}

In this section, we provide some further discussion, including the limitations of our proposed framework and future work.

First of all, we would like to highlight that the core of our proposed method is an existing public dataset for the target machine learning model our FL system tries to build collectively. The public dataset is considered validated and cannot be compromised by the attacker. While this assumption can be considered quite strong, it is reasonable for many real-world applications given the richness of public and open datasets in many fields. The public dataset effectively provides a ``ground truth'' benchmark so that we can train a reference model (RM) and an auditor model to test gradients (and updates) sent by all clients of the FL system. Note that such a public dataset does not need to stay static but can keep evolving by continuously accepting new and validated samples, such as how many open datasets are being maintained in many fields. Similarly, mistakenly added malicious samples can be removed from the public dataset. In a way, our proposed method leverages the community-wide review processes to maintain the integrity of the public dataset.

Although our proposed method has many merits, it also has some limitations. We discuss some of such limitations below, together with corresponding future research directions.

\begin{enumerate}
\item \textit{The need to maintain a stable and validated public dataset}: The dependence on a well-validated public dataset means that our method will not work if drastically different new data samples and even new classes keep emerging. Examples include automatic detection of rare diseases or rapidly changing viruses (e.g., COVID or flu viruses), and intrusion detection systems where attackers keep changing their attack behaviors. In other words, our proposed method is suitable only for applications where 1) class labels are stable and representative samples of each class do not change too rapidly, or 2) an active community can easily maintain a public dataset despite the fast evolution of data samples. In the future, we will study how to relax the dependency on the public dataset, which likely requires new ideas to update the \textbf{RM}, e.g., based on some distributed consensus protocols.

\item \textit{Dependence on a trusted third party or component}: Conceptually speaking, our proposed model relies on a trusted third party or a trusted component of the global server -- the auditor model + the reference model + the public dataset. Such dependencies are widely used in cryptography. In some applications, such a trusted party or model may be difficult to find or establish, e.g., when the clients do not have trust in each other and cannot agree on anything they can trust collectively. Whether it is possible to remove the trusted party/component or construct a similar entity under a ``zero trust'' environment is an interesting future research direction.

\item \textit{Sensitivity to ``small'' poison in updates}: The anomaly detection component of our proposed method still relies on a threshold to detect malicious clients, therefore it will not be able to detect ``small'' poisoning attacks. While in this case, a single attacker may find it difficult to poison the global model, carefully coordinated colluding attacks of many malicious clients may be able to collectively make a difference. Detecting such small colluding poisoning attacks requires different approaches and will be one interesting future research direction.

\item \textit{Limited experimental results}: Although we conducted many experiments, the existence of many poisoning attacks and different configurations of the FL system means that our experimental results do not cover all aspects of our proposed method. For instance, we used only three edge devices and one attacker and therefore did not consider different combinations of multiple attackers who could launch different attacks with different parameters, and some of them could collude. In the future, we plan to conduct a more comprehensive set of experiments to investigate more aspects of our proposed method's performance.

\item \textit{More advanced attacks}: Our experiments were conducted with known poisoning attacks that are unaware of our proposed method. Some advanced attacks may be developed to target our proposed method with a better evasion rate. Such advanced attacks and how our proposed method can be further improved will be another direction of future research.
\end{enumerate}

\section{Conclusions}
\label{sec:conclusions}

In this paper, we present a novel framework to detect poisoning attacks in FL applications. Our proposed framework can efficiently detect SOTA data and model poisoning attacks by observing the activations of the shared weights of the local models. Unlike most of the existing methods, our proposed framework can detect poisoning attacks without degrading the global model's performance. In addition, while most of the existing methods need knowledge about the number of attackers in the network, which can limit their applications, our proposed method can detect poisoning attacks without any knowledge about the number of attackers in the network. Moreover, in principle, our proposed method can detect any number of attackers, especially if they do not collude, thanks to the use of the auditor model based on the public dataset, while many SOTA methods can only work up to a particular number of attackers. Additionally, the time complexity of our proposed framework is dependent only on the number of edge devices, which makes it suitable to be used with any size (in terms of depth) of the network, whereas the time complexity of existing SOTA methods depends on the number of edge devices as well as the size of the network. We tested our proposed framework under four different data poisoning attacks and four different model poisoning attacks for two healthcare applications, showing that the proposed framework could efficiently detect malicious updates and exclude them from the global aggregation.

\section*{Acknowledgments}

This study was supported by a research project under the grant number LABEX/EQUIPEX, funded by the French National Research Agency under the frame program ``Investissements d'Avenir'' I-SITE ULNE / ANR-16-IDEX-0004 ULNE.

\bibliographystyle{IEEEtran}
\bibliography{main}

\end{document}